%% file: main.tex
\documentclass[sigconf,screen,nonacm]{acmart}

\renewcommand\footnotetextcopyrightpermission[1]{}
\usepackage{booktabs}
\usepackage{tabularx}
\usepackage{amsmath}
\usepackage{multirow}
\usepackage{xcolor}
\usepackage{graphicx}
\usepackage{subcaption}
\usepackage{microtype}
\usepackage{tikz}
\usetikzlibrary{positioning,arrows.meta,fit,backgrounds}

\newcommand{\CGCMAfix}{\textsc{CGCMA-fix}}   
\newcommand{\CGCMAlg}{\textsc{CGCMA-lg}}     
\newcommand{\SACMA}{\textsc{CGCMA-fix}}      
\newcommand{\DualBranch}{\textsc{CGCMA-db}}  
\newcommand{\CGCMAdb}{\textsc{CGCMA-db}}     
\newcommand{\CGCMA}{\textsc{CGCMA}}          
\newcommand{\dataset}{\textsc{CryptoMI}}

\begin{document}

\title{CGCMA: Conditionally-Gated Cross-Modal Attention for Event-Conditioned Asynchronous Fusion}

\author{Yunxiang Guo}
\affiliation{
  \institution{Independent Researcher}
  \country{China}
}
\email{guo_yx@outlook.com}

\begin{abstract}
We study \emph{asynchronous alignment}, a first-class multimodal learning setting in which a dense primary stream must be fused with sporadic external context whose value depends on when it arrives. Unlike standard multimodal benchmarks that assume structural synchrony, this setting requires models to reason explicitly about freshness and trust. We focus on the event-conditioned case in which continuous market states are paired with delayed web intelligence, and we use high-frequency cryptocurrency markets only as a timestamped, high-noise stress test for this broader problem. We propose \CGCMA{} (\textbf{C}onditionally-\textbf{G}ated \textbf{C}ross-\textbf{M}odal \textbf{A}ttention), whose central design principle is to separate text-conditioned grounding from lag-aware trust control. Text first attends over price sequences to identify event-relevant market states, after which a conditional gate uses modality agreement, web features, and lag $\tau_{\mathrm{lag}}$ to regulate residual injection and fall back toward unimodal prediction when external context is stale or contradictory. We introduce \dataset{} (\textbf{Crypto} \textbf{M}arket \textbf{I}ntelligence), an asynchronous evaluation corpus with 27{,}914 real-news samples pairing high-frequency price sequences with lagged web intelligence. On the current short real-news corpus, \CGCMA{} attains the highest mean downstream Sharpe ratio ($\mathbf{+0.449\pm0.257}$) among the evaluated baselines under a shared zero-cost threshold-trading evaluation on news-available bars. Additional controls show that the gain is not explained by web scalars alone and is not recovered by simple freshness heuristics. The resulting evidence supports problem validity and a promising asynchronous multimodal gain on this stress-test setting.
\end{abstract}



\maketitle

\input{sections/01_introduction}
\input{sections/02_related_work}
\input{sections/03_problem_formulation}
\input{sections/04_method}
\input{sections/05_experimental_protocol}
\input{sections/06_results}
\input{sections/08_limitations_ethics}
\input{sections/09_conclusion}

\bibliographystyle{ACM-Reference-Format}
\bibliography{references}

\end{document}

%% file: sections/01_introduction.tex
\section{Introduction}
\label{sec:intro}

Multimodal learning has matured on settings where modalities are structurally synchronized: video frames align with audio waveforms, captions are written with the image present, and transcripts are anchored to utterance boundaries~\cite{baltrusaitis2019multimodal,tsai2019multimodal}.
Real systems are often less forgiving.
External context can arrive late, sparsely, and with variable usefulness relative to a continuously evolving stream.
We refer to this challenge as \emph{asynchronous alignment}: the problem of fusing a dense primary signal with sporadic secondary context whose predictive value depends on when it arrives.
Despite its practical importance, asynchronous alignment remains underexplored because many standard benchmarks hide it through structural synchrony and because temporal validity is usually treated as metadata rather than as a modeling target~\cite{baltrusaitis2019multimodal,dasgupta2018hyte,liu2021temporal}.
In this paper we focus on one especially important subcase, \emph{event-conditioned asynchronous fusion}, where a dense primary stream must be updated using delayed textual context emitted by external events.
High-frequency cryptocurrency markets provide a particularly strict stress test for this problem: price time-series evolve under a dense forward-only clock, while web intelligence---news, alerts, and sentiment cascades---arrives intermittently with delays that can materially change its value.
A news item published 8 minutes after a liquidity shock carries very different signal than one published 3 hours later, after the market has largely adjusted.
This \emph{modality lag} ($\tau_{\mathrm{lag}}$) is therefore not noise to normalize away, but a first-class property of asynchronous multimodal prediction.

Existing multimodal architectures handle asynchronous signals inadequately~\cite{baltrusaitis2019multimodal,tsai2019multimodal}.
Methods requiring strict synchronization~\cite{sawhney2020multimodal,qin2019mdrm} discard most available intelligence, as modalities rarely co-occur within narrow windows.
Window-aggregation approaches~\cite{xu2018stocknet,ding2015deepevent} treat fresh and stale signals as interchangeable, conflating pre- versus post-event information.
Large language models~\cite{lopez2023fingpt,wu2023bloomberggpt} excel at language understanding but lack architectures for asynchronous alignment.
The result is a systematic blind spot: modality lag is rarely modeled as an architectural input.

We therefore frame asynchronous alignment as a core multimodal learning problem rather than as a finance-specific modeling trick.
We introduce \dataset{} (\textbf{Crypto} \textbf{M}arket \textbf{I}ntelligence), an asynchronous multimodal evaluation corpus that pairs high-frequency price sequences with lagged web intelligence across Bitcoin, Ethereum, and Solana.
\dataset{} records $\tau_{\mathrm{lag}}$ explicitly, enabling lag-conditioned architectures and stratified analysis.
The corpus includes a controlled synthetic development split (7{,}335 BTC samples) and a real-news extension from CryptoCompare (27{,}914 pooled BTC/ETH/SOL samples, December 2025--March 2026), letting us study asynchronous fusion in a real, high-noise, timestamped environment.
Cryptocurrency markets serve here only as a strict stress test: the same design questions arise whenever a dense primary modality must be updated using delayed, sporadic context, such as live media streams with delayed reactions or sensor streams with intermittent metadata.

We propose \CGCMA{}, a conditional cross-modal gating architecture for asynchronous prediction.
Text first attends over the full price sequence to produce grounded context, after which a per-dimension residual gate decides how much of that context to inject based on price--text agreement, web features, and $\tau_{\mathrm{lag}}$ (see Section~\ref{sec:method}).
The gate can close automatically when external context is stale or contradictory, degrading gracefully toward unimodal prediction.
We evaluate \CGCMA{} on \dataset{} against price-only, text-only, representative generic multimodal baselines~\cite{xu2018stocknet,tsai2019multimodal,zadeh2017tensor}, and task-specific lag-aware controls under a shared protocol.
The paper's central claim is therefore not that cryptocurrency forecasting is solved, but that event-conditioned asynchronous fusion can be operationalized as a concrete multimodal setting with measurable freshness structure, controlled evaluation, and a plausible architecture for lag-aware trust.

Our contributions are three-fold:
\begin{enumerate}
  \item \textbf{We define and instantiate an event-conditioned asynchronous fusion setting in which lag is part of both sample construction and evaluation.}
  \dataset{} exposes $\tau_{\mathrm{lag}}$ explicitly across synthetic and real-news corpora, enabling lag-stratified analysis and freshness-aware evaluation under a causal walk-forward protocol.
  The resulting corpus reveals a non-monotonic utility pattern, with the strongest web signal appearing in the 30--60 minute post-publication range.
  \item \textbf{We propose a design principle for asynchronous fusion that separates grounding from trust control, and we instantiate it in \CGCMA{}.}
  Text first attends over the dense price stream to identify event-relevant market states, after which a learned per-dimension gate decides how much of that grounded context should be injected based on agreement, web features, and lag.
  \item \textbf{We provide controlled evidence for grounded conditional fusion on the current real-news stress test.}
  Under a shared pooled protocol, \CGCMA{} achieves the strongest mean downstream Sharpe on the current real-news corpus ($+0.449\pm0.257$ across 8 runs), while targeted controls rule out web scalars alone and simple freshness heuristics as sufficient explanations.
  We position these results as supporting evidence for the asynchronous setting and the proposed design principle, rather than as a definitive claim of broad superiority.
\end{enumerate}

The remainder of this paper is organized as follows.
Section~\ref{sec:related} situates this paper in the literature.
Section~\ref{sec:problem} formalizes asynchronous multimodal market prediction.
Section~\ref{sec:method} describes the \CGCMA{} architecture and its asynchronous input structure.
Sections~\ref{sec:exp} and~\ref{sec:results} present the evaluation protocol and results.
Sections~\ref{sec:limitations}--\ref{sec:conclusion} discuss limitations, implications, and conclusions.

%% file: sections/02_related_work.tex
\section{Related Work}
\label{sec:related}

\paragraph{Financial forecasting from price and text.}
Sequence models over OHLCV data, including Transformer-based encoders~\cite{vaswani2017attention,zhou2021informer,wu2021autoformer}, temporal fusion models~\cite{lim2021tft}, and newer long-horizon backbones such as PatchTST, TimesNet, and iTransformer~\cite{nie2023patchtst,wu2023timesnet,liu2024itransformer}, remain strong baselines for short-horizon financial prediction, but they are structurally limited to endogenous market dynamics.
Text-augmented models extend this setting by incorporating exogenous signals from news, social media, or earnings calls.
StockNet~\cite{xu2018stocknet} jointly models price histories and Twitter text, while event-driven approaches such as Ding et al.~\cite{ding2015deepevent} show that structured news events can improve prediction.
More recent financial language models, including FinBERT~\cite{yang2020finbert}, FinGPT~\cite{lopez2023fingpt}, and BloombergGPT~\cite{wu2023bloomberggpt}, further improve text-side representation quality.
However, most of this literature assumes that text and price can be synchronously paired, either by aggregating text into fixed periods or by restricting to narrow publication windows.
That assumption is weak for cryptocurrency markets, where web intelligence arrives continuously and with highly variable latency.

\paragraph{Multimodal fusion and the asynchronous-evaluation gap.}
While the financial forecasting literature provides strong unimodal and text-augmented baselines, it largely sidesteps the asynchronous challenge that broader multimodal fusion research has only recently begun to surface.
More broadly, multimodal learning has been organized around representation, translation, alignment, fusion, and co-learning problems~\cite{baltrusaitis2019multimodal}, with gating-based and memory-based fusion families such as GMU, MFN, MISA, and DMNC~\cite{arevalo2017gmu,zadeh2018mfn,hazarika2020misa,le2018dmnc} highlighting the importance of selective cross-modal interaction over time and MulT~\cite{tsai2019multimodal} making the unaligned-sequence setting especially visible for language-centered benchmarks.
Multimodal financial systems such as MDRM~\cite{qin2019mdrm} and the earnings-based risk framework of Sawhney et al.~\cite{sawhney2020multimodal} demonstrate that joint modeling across modalities can improve downstream performance, but they operate in settings with tightly scheduled events and much easier synchronization.
General-purpose fusion architectures therefore provide the main methodological reference points for our comparisons: BiLSTM late fusion~\cite{xu2018stocknet}, directional cross-modal attention in MulT~\cite{tsai2019multimodal}, multiplicative interaction modeling in TFN~\cite{zadeh2017tensor}, modality gating in GMU~\cite{arevalo2017gmu}, memory-based sequential fusion in MFN~\cite{zadeh2018mfn}, and the broader fusion taxonomy summarized by Baltru\v{s}aitis et al.~\cite{baltrusaitis2019multimodal}.
In cryptocurrency prediction, prior multimodal datasets and systems remain mostly daily and do not expose within-day freshness.
Sentiment from Twitter~\cite{abraham2018cryptocurrency}, Reddit~\cite{kim2019predicting}, fear\,\&\,greed indicators~\cite{chen2021fear}, and on-chain activity~\cite{kondor2014rich} has been studied separately, while integrated benchmarks such as PreBit~\cite{zou2023prebit}, DAM~\cite{li2024dam}, CryptoPulse~\cite{liu2024cryptopulse}, and Barth{\'e}lemy et al.~\cite{barthelemy2025deepnlp} operate at daily granularity.
As summarized in Table~\ref{tab:benchmark_comparison}, these datasets do not record modality lag $\tau_{\mathrm{lag}}$ per aligned sample, which makes lag-conditioned training and lag-stratified analysis impossible.
\dataset{} is designed to fill this broader asynchronous-evaluation gap by providing 15-minute price alignment together with explicit lag metadata; cryptocurrency markets serve here as a convenient stress-test environment rather than as the paper's only intended downstream domain.

\begin{table*}[t]
\caption{Comparison of cryptocurrency multimodal datasets. \emph{Lag $\tau$} indicates whether per-sample modality lag is recorded; \emph{Rolling+SR} indicates rolling walk-forward evaluation with Sharpe reporting. LLM synthesis trades source authenticity for temporal control and sub-hourly coverage; the associated limitations are discussed in Section~\ref{sec:limitations}.}
\label{tab:benchmark_comparison}
\centering
\small
\setlength{\tabcolsep}{4pt}
\renewcommand{\arraystretch}{1.05}
\begin{tabular}{llccccc}
\toprule
Dataset & Text source & Price granularity & Assets & Samples & Lag $\tau$ recorded & Rolling+SR \\
\midrule
PreBit~\cite{zou2023prebit}            & Real (Twitter)          & Daily   & BTC         & ${\sim}$1K         & \texttimes & \texttimes \\
DAM~\cite{li2024dam}                   & Real (CryptoCompare)    & Daily   & BTC         & ${\sim}$1K         & \texttimes & \texttimes \\
CryptoPulse~\cite{liu2024cryptopulse}  & Real (Cointelegraph)    & Daily   & Top-5       & ${\sim}$1K         & \texttimes & \texttimes \\
Barth{\'e}lemy et al.~\cite{barthelemy2025deepnlp} & Real (Twitter/Reddit) & Daily & BTC/ETH  & ${\sim}$3yr        & \texttimes & \checkmark \\
\midrule
\dataset{} (synthetic)                & LLM-synthesized         & \textbf{15 min} & BTC/ETH/SOL & \textbf{21{,}374} & \checkmark & \checkmark \\
\dataset{} (real news)                & Real (CryptoCompare)    & \textbf{15 min} & BTC/ETH/SOL & \textbf{27{,}914} & \checkmark & \checkmark \\
\bottomrule
\end{tabular}
\end{table*}

\paragraph{Temporal alignment, freshness, and lag-aware trust.}
Beyond finance, related ideas appear in temporal knowledge graph embedding, temporally weighted recommendation, and sequence models that explicitly acknowledge unaligned context windows.
HyTE~\cite{dasgupta2018hyte} conditions representations on validity intervals, temporally weighted co-attention methods in recommendation down-weight stale context~\cite{liu2021temporal}, and MulT-style cross-modal attention shows that alignment need not be hard-coded at equal timestamps~\cite{tsai2019multimodal}.
However, these lines of work stop short of a learned, content-aware trust mechanism that jointly conditions on cross-modal agreement and lag.
What is missing in multimodal financial forecasting is therefore not just a better text encoder, but an explicit mechanism for deciding \emph{when} text should be trusted relative to the current market state.
Existing financial fusion models typically ignore modality lag or treat it as a nuisance variable rather than as an architectural input.
\CGCMA{} takes the opposite view: text first grounds into the price history through cross-modal attention, and lag-aware trust is then handled explicitly by a conditional gate that receives $\tau_{\mathrm{lag}}$ together with the modality representations.

%% file: sections/03_problem_formulation.tex
\section{Problem Formulation}
\label{sec:problem}

\subsection{Asynchronous Multimodal Market Prediction}

Let $\mathcal{P}_t = \{p_{t-L+1}, \ldots, p_t\}$ denote a price sequence of $L$ OHLCV bars ending at time $t$, where $p_i \in \mathbb{R}^5$ encodes open, high, low, close, and volume.

Let $\mathcal{W}_{t^*}$ denote a web intelligence snapshot collected at time $t^* \leq t$, aggregating news summaries, fear\,\&\,greed index, on-chain whale signals, and social sentiment into a structured representation.
The \emph{modality lag} is defined as $\tau_{\mathrm{lag}} = t - t^*$, measured in minutes.

\paragraph{Prediction target.}
We formulate a binary classification problem over a horizon of $H$ bars:
\[
y_t = \mathbf{1}\!\left[\frac{c_{t+H} - c_t}{c_t} > 0\right],
\]
where $c_t$ denotes the close price at bar $t$.

\paragraph{Asynchronous alignment constraint.}
Web intelligence is available only at event times $t^* \in \mathcal{E}$, where $\mathcal{E}$ is an irregular discrete set.
Given a price bar at time $t$, we select the most recent event $t^* = \max\{e \in \mathcal{E} : e \leq t\}$.
A sample $(t, t^*)$ is retained only if $\tau_{\mathrm{lag}} = t - t^* \leq \tau_{\max}$ (maximum allowed lag).
This formulation explicitly captures the practical reality that web intelligence may be stale relative to the current price state.

\paragraph{Objective.}
Given a dataset $\mathcal{D} = \{(\mathcal{P}_t,\, \mathcal{W}_{t^*},\, \tau_{\mathrm{lag}},\, y_t)\}$, learn a model $f: (\mathcal{P}_t, \mathcal{W}_{t^*}, \tau_{\mathrm{lag}}) \to [0,1]$ that estimates $P(y_t = 1)$ while remaining robust to changes in assets and lag distributions.
Evaluation uses the rolling walk-forward protocol in Section~\ref{sec:exp}, with AUC and downstream Sharpe as the primary metrics.

%% file: sections/04_method.tex
\section{Dataset and Method}
\label{sec:method}

\subsection{Task Inputs and Alignment}
Each decision uses a 15-minute OHLCV lookback window with $L=64$ bars, a web-intelligence text blob, and 13 aligned scalar web features.
At decision time $t$, we pair the price window with the most recent web item $t^* \leq t$ such that $\tau_{\mathrm{lag}} = t - t^* \leq 180$ minutes.
Figure~\ref{fig:arch} denotes the resulting causally aligned text signal as an \emph{aligned text context}.
Detailed corpus statistics and evaluation-split properties are deferred to Section~\ref{sec:exp} and Table~\ref{tab:dataset}.

\subsection{\CGCMA{}: Conditionally-Gated Cross-Modal Attention}
\label{sec:cgcma}

\paragraph{Overview.}
\CGCMA{} consists of a \emph{price encoder}, a \emph{text encoder}, a \emph{text-conditioned price attention} module, and a \emph{conditional gate}, as illustrated in Figure~\ref{fig:arch}.
The central design principle is to separate \emph{grounding} from \emph{trust control}: attention first uses event semantics to retrieve relevant market states from the price history, and a learned gate then determines how strongly that text-conditioned signal should update the primary price representation.
Viewed more abstractly, event-conditioned asynchronous fusion contains two coupled subproblems.
The \emph{grounding problem} asks which parts of a dense primary stream are relevant to the current external context.
The \emph{trust problem} asks how much of that context should be injected, given freshness, agreement, and auxiliary quality signals.
\CGCMA{} maps these two subproblems to two separate modules rather than collapsing them into a single fusion step.

\begin{figure*}[t]
  \centering
  \includegraphics[width=0.95\textwidth,keepaspectratio]{figures/cgcma_arch_v2.drawio.png}
  \Description{Architecture of CGCMA. All inputs are causally aligned at decision time t: the OHLCV window ends at t, the aligned text context satisfies t star less than or equal to t, the web scalars are available at t, and the freshness variable is tau equals t minus t star. A price encoder outputs the full price token sequence H superscript p and a price summary h sub p. A text encoder outputs h sub t, and a web scalar encoder outputs h sub w. Text-conditioned price attention uses the text token as query and the price token sequence as keys and values, followed by layer normalization, to retrieve an event-conditioned market context h sub c. A conditional gate receives h sub p, h sub c, their context-shift term delta sub pc equals h sub p minus h sub c, h sub w, and tau divided by 60, and outputs a per-dimension gate vector g. A gated residual fusion block computes h sub f equals h sub p plus g elementwise-multiplied by h sub c, preserving a price-only fallback path when the gate closes. The fused state is concatenated with a direct predictive web cue and passed to a task head, shown here as binary direction prediction.}
  \caption{\CGCMA{} architecture. The proposed fusion is explicitly split into two roles: text-conditioned price attention \emph{grounds} the aligned text in the full price sequence to form $\mathbf{h}_c=\mathrm{LN}(\mathrm{MHA}(\mathbf{h}_t,\mathbf{H}^p,\mathbf{H}^p))$, while the conditional gate \emph{controls trust} in that context using the context shift $\mathbf{\Delta}_{pc}=\mathbf{h}_p-\mathbf{h}_c$, web context, and freshness $\tau/60$. This yields a causally aligned multimodal residual update $\mathbf{h}_f=\mathbf{h}_p+\mathbf{g}\odot\mathbf{h}_c$ that falls back to the price-only path when text is stale or uninformative, while retaining $\mathbf{h}_w$ as a direct predictive cue at the task head. The main experiments instantiate the task head as binary direction classification, but the fusion design itself is task-agnostic.}
  \label{fig:arch}
\end{figure*}

\paragraph{Price encoder.}
The price sequence $\mathcal{P}_t \in \mathbb{R}^{L \times d_p}$ is projected to model dimension $d$, augmented with learnable positional embeddings, and encoded by a $N_p$-layer Transformer with a prepended \textsc{[CLS]} token:
\[
[\,\mathbf{h}_{p},\, \mathbf{H}^p\,] = \mathrm{TransformerEncoder}([\mathbf{e}_{\mathrm{cls}},\; \mathbf{W}_p \mathcal{P}_t + \mathbf{E}_{\mathrm{pos}}]),
\]
where $\mathbf{h}_{p} \in \mathbb{R}^d$ is the global price summary and $\mathbf{H}^p \in \mathbb{R}^{L \times d}$ is the full price token sequence.

\paragraph{Text and web encoder.}
The web intelligence text blob is encoded with a frozen multilingual sentence transformer~\cite{reimers2019sentencebert} (paraphrase-multilingual-MiniLM-L12-v2, 384 dimensions), producing $\mathbf{e}^{\mathrm{text}} \in \mathbb{R}^{384}$.
The 13 scalar web features are normalized into $\mathbf{w} \in \mathbb{R}^{13}$.
The text branch is projected to model width $d$, while the web branch is projected to a separate width $d_w$ (implemented as half of the gate hidden width):
\[
\mathbf{h}_{t} = \mathrm{LN}(\mathbf{W}_{\mathrm{text}}\,\mathbf{e}^{\mathrm{text}}),\quad
\mathbf{h}_{w} = \mathrm{MLP}_{\mathrm{web}}(\mathbf{w}),
\]
where $\mathbf{h}_{t} \in \mathbb{R}^{d}$ and $\mathbf{h}_{w} \in \mathbb{R}^{d_w}$.

\paragraph{Text-conditioned price attention.}
The text token attends over the full price token sequence to extract price-grounded context:
\begin{equation}
\tilde{\mathbf{h}}_{t \to p} = \mathrm{MHA}(\mathbf{h}_{t},\; \mathbf{H}^p,\; \mathbf{H}^p),
\label{eq:text2price}
\end{equation}
where $\mathbf{h}_{t}$ serves as the query, and $\mathbf{H}^p$ provides keys and values.
This stage addresses the \emph{grounding} problem: using the text representation as the query lets current event semantics retrieve the most relevant market states from the historical price trajectory, rather than forcing the full price sequence to absorb the text signal uniformly.
In the current implementation, the attention output is layer-normalized directly, i.e.,
\[
\mathbf{h}_c = \mathrm{LN}(\tilde{\mathbf{h}}_{t \to p}),
\]
yielding an event-conditioned market context grounded in the current price dynamics.

\paragraph{Conditional gate.}
Unlike fixed-schedule staleness gating (e.g., exponential decay), \CGCMA{} uses a \emph{learned} per-dimension gate that simultaneously models cross-modal agreement, web signal quality, and modality lag.
This stage addresses the \emph{trust} problem by deciding how much of the grounded text-conditioned context should influence the price representation:
\begin{equation}
\mathbf{\Delta}_{pc} = \mathbf{h}_{p} - \mathbf{h}_c,\qquad
\mathbf{g} = \sigma\!\left(W_g\,[\mathbf{h}_{p},\; \mathbf{h}_c,\; \mathbf{\Delta}_{pc},\; \mathbf{h}_{w},\; \tau_{\mathrm{lag}} / 60]\right) \in (0,1)^d,
\label{eq:cgcma_gate}
\end{equation}
where the context-shift term $\mathbf{\Delta}_{pc}$ measures the signed deviation of the event-conditioned market context from the price-only summary, and $\tau_{\mathrm{lag}} = t - t^\ast$ denotes the age of the aligned text context at decision time $t$; we scale it by $60$ so the gate receives freshness in hours.
The gate operates \emph{per-dimension}, enabling selective feature-level fusion: individual dimensions of the text context can be gated in or out independently, providing finer-grained control than a scalar gate.

\paragraph{Gated residual fusion and prediction head.}
The text context is added to the price representation as a learned residual:
\begin{equation}
\mathbf{h}_{f} = \mathbf{h}_{p} + \mathbf{g} \odot \mathbf{h}_c,
\label{eq:cgcma_fusion}
\end{equation}
where $\odot$ denotes element-wise multiplication.
When $\mathbf{g} \to \mathbf{0}$---which occurs when text is stale, uninformative, or contradicts the price signal---the model degrades gracefully to the price-only summary representation $\mathbf{h}_p$.
The fused representation and web features are concatenated and passed through a two-layer GELU MLP. In this design, $\mathbf{h}_w$ plays two distinct roles: it conditions the gate as a context-quality signal and is also preserved as a direct predictive feature at the task head:
\[
\hat{y} = \mathrm{MLP}(\mathrm{LN}([\mathbf{h}_{f},\; \mathbf{h}_{w}])).
\]
Accordingly, the concatenated task-head input has dimension $d + d_w$ rather than $2d$.

\paragraph{Comparison with fixed-gate staleness control.}
\CGCMA{} differs from the fixed-gate predecessor \SACMA{} in two key respects.
First, the staleness gate in \SACMA{} is a scalar exponential function of $\tau_{\mathrm{lag}}$ alone ($\exp(-\lambda \tau / 60)$), applying a uniform decay to all text dimensions regardless of content.
\CGCMA{}'s gate is a learned MLP operating on the full context vector (price embedding, text--price context, their difference, web signals, and lag), enabling content-aware and lag-aware gating simultaneously.
Second, \CGCMA{} uses a single cross-attention pass (text attends to price), whereas \SACMA{} uses bidirectional attention followed by a 3-token fusion Transformer.
The single-direction design reduces the parameter count and avoids the overfitting risk of bidirectional attention at low sample counts.

\subsection{Baselines}
\label{sec:baselines}

We compare \CGCMA{} against protocol baselines and representative published multimodal fusion architectures:

\begin{itemize}
  \item \textbf{Price-only Transformer} (\textsc{PriceTx-S})~\cite{vaswani2017attention}: Transformer over the price sequence with a CLS-based prediction head; primary unimodal baseline.
  \item \textbf{Text-only encoder} (\textsc{TextOnly})~\cite{reimers2019sentencebert}: frozen multilingual MiniLM embedding with an MLP head.
  \item \textbf{Linear TF-IDF Late Fusion} (\textsc{TFLate})~\cite{hoerl1970ridge,salton1988term}: ridge price and TF-IDF text models combined by validation-selected late-fusion weighting; used only in the synthetic subset study.
  \item \textbf{BiLSTM Fusion} (\textsc{BiLSTM})~\cite{xu2018stocknet}: bidirectional LSTM price encoder fused with text and web projections by concatenation.
  \item \textbf{Early Fusion} (\textsc{EarlyFusion})~\cite{baltrusaitis2019multimodal}: concatenation of price, text, and web features before a shared MLP head.
  \item \textbf{MulT} (\textsc{MulT})~\cite{tsai2019multimodal}: bidirectional cross-modal attention with pooled classification.
  \item \textbf{Tensor Fusion Network} (\textsc{TFN})~\cite{zadeh2017tensor}: tensor outer-product fusion of unimodal subspaces followed by an MLP head.
\end{itemize}

In addition, the development-path analysis retains \textbf{Dual-Branch Conditional Late Fusion} (\DualBranch{}), an internal ablation predecessor with dense semantic and sparse lexical text branches fused through lag-aware residual gates.

All models use $d{=}32$, dropout 0.3, AdamW (lr $10^{-3}$, wd $10^{-3}$), batch 64, 30 epochs with patience 7, and AUC-based checkpoint selection.
PriceTx-S, TextOnly, EarlyFusion, and TFLate are protocol baselines instantiated within one shared evaluation framework; BiLSTM, MulT, and TFN are adaptations of named published architectures.
Primary real-news results (Table~\ref{tab:main}) compare \textsc{PriceTx-S}, \textsc{TextOnly}, \textsc{EarlyFusion}, \textsc{BiLSTM}, \textsc{MulT}, \textsc{TFN}, and \CGCMA{} under one shared protocol.
\DualBranch{} is retained as an ablation predecessor within the development path rather than the final real-news comparison table.

%% file: sections/05_experimental_protocol.tex
\section{Experimental Protocol}
\label{sec:exp}

\subsection{\dataset{} Summary and Task Support}
We collect 15-minute OHLCV data from Binance Futures for BTCUSDT, ETHUSDT, and SOLUSDT over January 2025--March 2026.
Each sample uses a lookback window of $L=64$ bars (16 hours), per-window $z$-score normalization, and eleven derived technical features.

\dataset{} contains two complementary corpora.
The \emph{synthetic corpus} aggregates LLM-synthesized news, fear\,\&\,greed, on-chain, and social signals into structured JSON snapshots with 13 scalar fields and a text blob; January--September 2025 is almost entirely bullish ($>$99\% positive).
The \emph{real-news corpus} is built from CryptoCompare articles~\cite{cryptocompare2025} over December 2025--March 2026 and forms the paper's primary evaluation set: 27{,}914 pooled samples with much lower mean lag and substantially more diverse sentiment.

For each 15-minute price bar at time $t$, we select the most recent web item $t^* \leq t$ with $\tau_{\mathrm{lag}} = t - t^* \leq 180$ minutes.
Bars without a valid web context are excluded from the multimodal sample set.
This defines a conditional forecasting task on news-available bars, and the price-only baseline in the main comparison is evaluated on that same filtered support.

\begin{table}[t]
\caption{\dataset{} statistics. \emph{Synthetic} denotes the LLM-synthesized corpus from Jan--Dec 2025; \emph{Real news} denotes the CryptoCompare corpus from Dec 2025--Mar 2026. We report sample count, mean lag, and positivity rate.}
\label{tab:dataset}
\centering
\small
\setlength{\tabcolsep}{4pt}
\renewcommand{\arraystretch}{1.05}
\begin{tabular}{llrrr}
\toprule
Corpus & Asset & Samples & Lag (min) & Pos. rate \\
\midrule
\multirow{3}{*}{Synthetic} & BTC & 7{,}335 & 62.3 & $>$99\% \\
 & ETH & 7{,}039 & 64.1 & $>$99\% \\
 & SOL & 7{,}000 & 63.8 & $>$99\% \\
\midrule
\multirow{3}{*}{Real news} & BTC & 19{,}241 & 5.9 & 45\% \\
 & ETH & 6{,}349 & 6.1 & 45\% \\
 & SOL & 2{,}333 & 5.7 & 44\% \\
\bottomrule
\end{tabular}
\end{table}

\subsection{Rolling Walk-Forward Splits}
We adopt a rolling walk-forward evaluation to prevent any look-ahead bias.
The \dataset{} samples are ordered chronologically and partitioned into windows of 15-minute bars.
We use three rolling protocols depending on the dataset scale and evaluation goal.
\emph{Standard protocol} (ablation experiments): minimum training windows $K_{\mathrm{train}}=40$, validation $K_{\mathrm{val}}=16$, test $K_{\mathrm{test}}=12$, step $K_{\mathrm{step}}=8$, yielding \textbf{111 folds} for BTC (1{,}136 filtered samples, min\_strength${=}0.2$), ${\approx}38$--111 folds for other assets.
\emph{Scaling protocol} (synthetic full-year study): $K_{\mathrm{train}}=500$, $K_{\mathrm{val}}=200$, $K_{\mathrm{test}}=200$, $K_{\mathrm{step}}=100$, yielding \textbf{63 folds} for BTC (7{,}335 samples) with ${\approx}500$ training samples per fold.
\emph{Non-overlapping protocol} (real-news validation): $K_{\mathrm{train}}=500$, $K_{\mathrm{val}}=200$, $K_{\mathrm{test}}=200$, $K_{\mathrm{step}}=200$ (step equals test size, so test windows are non-overlapping), yielding about 35 pooled folds on the real-news corpus (32--35 depending on asset/filtering).
The non-overlapping protocol eliminates temporal correlation between adjacent test folds, providing the most conservative multi-fold Sharpe estimates.
For each fold:
(1) models are trained from scratch on the training split,
(2) early stopping is performed on the validation split,
(3) predictions are made on the test split without any re-fitting.

\subsection{Downstream Trading Evaluation}
Beyond classification metrics, we evaluate the practical utility of model outputs through a threshold-based trading strategy:
\begin{itemize}
  \item \textbf{Long} if $\hat{p}_t \geq \theta_{\mathrm{long}} = 0.55$;
  \item \textbf{Short} if $\hat{p}_t \leq \theta_{\mathrm{short}} = 0.45$;
  \item \textbf{Flat} otherwise.
\end{itemize}
For each trade at time $t$ with horizon $H$, the realized return is $r_t = \mathrm{sign}(\mathrm{position}) \cdot \frac{c_{t+H} - c_t}{c_t}$.
The downstream Sharpe ratio is computed as $\mathrm{SR} = \frac{\mathbb{E}[r]}{\mathrm{std}(r)} \cdot \sqrt{N}$ where $N$ is the number of test windows.
This metric is used as a decision-oriented proxy for downstream utility rather than as a full live-trading simulation.
All reported downstream trading metrics assume zero transaction costs and zero slippage; they should therefore be interpreted as controlled offline comparisons rather than live-trading projections.

\subsection{Metrics}
\begin{itemize}
  \item \textbf{AUC}: area under the ROC curve for binary price-direction classification.
  \item \textbf{Brier score}: mean squared probability error, measuring calibration.
  \item \textbf{Downstream Sharpe}: risk-adjusted return of the threshold strategy applied to model outputs.
  \item \textbf{Max drawdown}: worst peak-to-trough equity decline over test folds.
  \item \textbf{Hit rate}: fraction of trades with positive realized return.
\end{itemize}
We report mean $\pm$ standard deviation across rolling folds.
For the dedicated control sweeps (Tables~\ref{tab:design} and~\ref{tab:v5_scaling}), we report mean $\pm$ std across \textbf{4 independent runs} (different random seeds); single-run rankings at small-scale folds are unreliable.
For the final pooled real-news comparison (Table~\ref{tab:main}), we report mean $\pm$ std across \textbf{8 independent runs}, which provides stronger stability evidence while still stopping short of a broad claim of definitive statistical separation.
For that main real-news comparison, we additionally report run-level 95\% $t$-intervals over the eight seed means and matched-fold bootstrap intervals for key pairwise Sharpe deltas using the shared fold assignments.

%% file: sections/06_results.tex
\section{Results}
\label{sec:results}

We present the empirical results as a staged evidence chain rather than as a single leaderboard claim.
The guiding questions are:
\emph{(Q1)} does modality lag create a meaningful asynchronous fusion problem at all;
\emph{(Q2)} are simple freshness heuristics sufficient once lag is observed;
\emph{(Q3)} which parts of the final design receive direct support from controlled sweeps; and
\emph{(Q4)} what is the strongest quantitative conclusion we can defend on the current short real-news corpus.
Table~\ref{tab:main} addresses the fourth question directly, while the later subsections use lag analysis, task-specific controls, and dedicated ablations to build a supporting mechanism-oriented argument around that main comparison.

\subsection{Main Comparison: Real-News Multi-Asset Evaluation}
\label{sec:main}

Table~\ref{tab:main} reports the primary comparison on the \dataset{} real-news corpus (BTC+ETH+SOL pooled, 27{,}914 samples) under non-overlapping rolling walk-forward evaluation ($K_{\mathrm{step}}{=}K_{\mathrm{test}}{=}200$, ${\approx}35$ folds).
Results are mean$\pm$std across 8 independent seeds.

\begin{table*}[t]
\caption{Main real-news comparison on \dataset{} (BTC+ETH+SOL pooled; 27{,}914 samples; non-overlapping rolling evaluation; ${\approx}35$ folds). Entries are mean $\pm$ std across \textbf{8 runs}. Best mean is \textbf{bold}; second best is \underline{underlined}.}
\label{tab:main}
\centering
\small
\setlength{\tabcolsep}{5pt}
\renewcommand{\arraystretch}{1.14}
\begin{tabular}{@{} l @{\hspace{24pt}} c @{\hspace{10pt}} l @{\hspace{8pt}} r r c r r c @{}}
\toprule
Model & Inputs & Fusion & Mean $\uparrow$ & Std $\downarrow$ & 95\% CI & $\Delta$ vs.\ Price & $\Delta$ vs.\ BiLSTM & Hit rate $\uparrow$ \\
\midrule
\multicolumn{9}{@{}l}{\emph{Unimodal baselines}} \\
\textsc{PriceTx-S}  & P   & Transformer + CLS & $+0.194$ & $0.229$ & $[0.003,\,0.385]$ & --- & $-0.015$ & 0.518 \\
\textsc{TextOnly}   & T   & SBERT + MLP & $-0.043$ & $0.168$ & $[-0.183,\,0.098]$ & $-0.237$ & $-0.252$ & 0.495 \\
\addlinespace[2pt]
\midrule
\multicolumn{9}{@{}l}{\emph{Multimodal baselines}} \\
\textsc{PriceWeb}   & P+W & Price+web late fusion & $+0.161$ & $0.289$ & $[-0.299,\,0.621]$ & $+0.060$ & $-0.099$ & 0.532 \\
\textsc{EarlyFusion} & P+T+W & Early concat & $+0.198$ & $0.150$ & $[0.072,\,0.324]$ & $+0.004$ & $-0.011$ & 0.525 \\
\textsc{BiLSTM}     & P+T+W & BiLSTM late fusion & \underline{$+0.209$} & $0.196$ & $[0.045,\,0.372]$ & \underline{$+0.015$} & --- & 0.521 \\
\textsc{MulT}       & P+T+W & Cross-attn & $+0.187$ & $0.220$ & $[0.003,\,0.371]$ & $-0.007$ & $-0.022$ & 0.525 \\
\textsc{TFN}        & P+T+W & Tensor fusion & $-0.033$ & $0.118$ & $[-0.132,\,0.066]$ & $-0.227$ & $-0.242$ & 0.512 \\
\addlinespace[2pt]
\midrule
\multicolumn{9}{@{}l}{\emph{Proposed}} \\
\CGCMA{}            & P+T+W & Grounding + cond.\ gate & \textbf{$+0.449$} & $0.257$ & $[0.235,\,0.664]$ & \textbf{$+0.256$} & \textbf{$+0.241$} & \textbf{0.535} \\
\bottomrule
\end{tabular}
\vspace{2pt}
\parbox{0.90\textwidth}{\footnotesize \emph{Note.} P = price, T = text, W = web scalars. Hit rate is the proportion of executed trades with positive realized return under the shared threshold rule. All models use the same $d{=}32$ budget and rolling folds; all multimodal baselines receive $\tau_{\mathrm{lag}}$ as an explicit input for fairness.}
\end{table*}

Table~\ref{tab:main} is the paper's primary quantitative evidence and should be read as a controlled study of event-conditioned asynchronous fusion on the current short real-news corpus and on the restricted task support of news-available bars.
Within that setting, \CGCMA{} achieves the highest mean downstream Sharpe ($+0.449\pm0.257$) under the shared zero-cost threshold-trading evaluation, with a mean improvement of $\Delta{=}{+}0.256$ over the price-only baseline and $\Delta{=}{+}0.241$ over the strongest generic multimodal baseline (\textsc{BiLSTM}).
Using the shared fold assignments and averaging each fold's Sharpe across the eight seeds, the matched-fold bootstrap interval for \CGCMA{} minus \textsc{PriceTx-S} is $\Delta{=}{+}0.256$ with 95\% CI $[0.003,\;0.491]$, and 22 of 35 fold indices are positive.
This still falls short of a broad superiority claim, but it strengthens the evidence from favorable ranking to a more stable positive margin over the price-only baseline under the current protocol.

The rest of Table~\ref{tab:main} is best understood as a sequence of controls around that main result.
Among the generic multimodal baselines, \textsc{BiLSTM} is strongest at $+0.209\pm0.196$, followed closely by \textsc{EarlyFusion} at $+0.198\pm0.150$.
\textsc{PriceWeb}, which removes free-form text and keeps only price plus web scalars, reaches $+0.161\pm0.289$.
This is directionally better than \textsc{PriceTx-S} but still well below \CGCMA{}, indicating that the main gain is not explained by structured web scalars alone.
\textsc{MulT} remains near parity with price-only while exhibiting comparable variance, and both \textsc{TFN} and \textsc{TextOnly} underperform the price-only baseline.

At the classification level, the strongest models remain tightly clustered around AUC $\approx 0.53$ on the same real-news folds.
We therefore interpret \CGCMA{}'s advantage primarily as stronger downstream decision utility under the shared threshold-trading rule, rather than as a large jump in raw ranking accuracy.
That distinction is important for scope: the present evidence supports a promising asynchronous multimodal gain on the current corpus and task definition, not yet a definitive or market-wide superiority claim.
The remaining subsections answer narrower mechanism questions around this main result: whether lag creates an interpretable freshness structure, whether simple freshness heuristics are sufficient, and which parts of the final architecture receive direct support from dedicated sweeps.

\begin{table}[tbp]
\caption{Auxiliary predictive metrics for selected models from Table~\ref{tab:main}. Entries are 8-run means on the same pooled real-news protocol, except \textsc{PriceWeb}, which remains a 4-run control.}
\label{tab:main_aux}
\centering
\footnotesize
\setlength{\tabcolsep}{4pt}
\renewcommand{\arraystretch}{1.05}
\begin{tabular}{lccc}
\toprule
Model & AUC $\uparrow$ & Brier $\downarrow$ & Hit rate $\uparrow$ \\
\midrule
\textsc{PriceTx-S} & \textbf{0.537} & \textbf{0.261} & 0.518 \\
\textsc{PriceWeb} & 0.536 & \textbf{0.259} & 0.532 \\
\textsc{BiLSTM} & 0.528 & 0.263 & 0.521 \\
\CGCMA{} & 0.536 & 0.265 & \textbf{0.535} \\
\bottomrule
\end{tabular}
\vspace{2pt}
\parbox{0.98\columnwidth}{\footnotesize \emph{Note.} These metrics contextualize the Sharpe results. \textsc{PriceWeb} improves raw directional quality slightly over price-only, but does not reproduce the downstream utility gain of \CGCMA{}.}
\end{table}

\subsection{Real-News Task-Specific Control Comparison}
\label{sec:design}

To complement the generic multimodal baselines in Table~\ref{tab:main}, Table~\ref{tab:design} reports a separate dedicated four-seed sweep of task-specific lag-aware control variants under the same pooled real-news protocol.
All rows use the same BTC+ETH+SOL corpus and the same non-overlapping rolling geometry, but they come from a distinct run batch from the main 8-seed comparison.
Accordingly, Table~\ref{tab:design} is intended for \emph{within-sweep} comparison only: it tests whether simpler freshness heuristics can recover the gain without the full grounded conditional fusion used in \CGCMA{}, but its absolute Sharpe values should not be read as direct replacements for the 8-seed estimates in Table~\ref{tab:main}.

\begin{table}[tbp]
\caption{Dedicated four-seed sweep of task-specific lag-aware controls on the pooled real-news protocol. Entries are 4-run mean$\pm$std Sharpe from a run batch distinct from Table~\ref{tab:main}; compare rows within this table only.}
\label{tab:design}
\centering
\footnotesize
\setlength{\tabcolsep}{4pt}
\renewcommand{\arraystretch}{1.05}
\begin{tabular}{l p{0.34\columnwidth} c c}
\toprule
Model & Key design choice & Mean Sharpe & $\Delta$ vs.\ PriceTx-S \\
\midrule
\textsc{PriceTx-S} & price-only Transformer & $+0.101{\pm}0.197$ & --- \\
\textsc{FixedDecay} & fixed exponential decay & $+0.219{\pm}0.382$ & $+0.118$ \\
\textsc{LearnedDecay} & learned scalar decay & $+0.014{\pm}0.203$ & $-0.087$ \\
\textsc{HardFilter} & hard stale-text filter & $+0.272{\pm}0.408$ & $+0.171$ \\
\CGCMAdb{} & conditional late-fusion predecessor & $+0.244{\pm}0.185$ & $+0.143$ \\
\CGCMA{} & grounding + vector gate & $+0.375{\pm}0.270$ & $+0.274$ \\
\bottomrule
\end{tabular}
\vspace{2pt}
\parbox{0.98\columnwidth}{\footnotesize \emph{Note.} This sweep shares the same data support and rolling geometry as the main study, but comes from a separate run batch. Use Table~\ref{tab:main} for the headline absolute comparison.}
\end{table}

This table should be read for its within-sweep ordering rather than for absolute comparison with Table~\ref{tab:main}. Within that sweep, freshness-aware heuristics do help: \textsc{FixedDecay}, \textsc{HardFilter}, and \CGCMAdb{} all improve on \textsc{PriceTx-S}, which is consistent with modality lag carrying useful information rather than acting as pure nuisance. At the same time, those simpler controls do not recover the full gain of \CGCMA{}: the strongest of them, \textsc{HardFilter}, reaches $+0.272{\pm}0.408$, still below \CGCMA{}'s $+0.375{\pm}0.270$. \textsc{LearnedDecay} remains near parity at $+0.014{\pm}0.203$, suggesting that scalar staleness weighting alone is unstable at this evaluation scale. Taken together, these sweeps support the view that lag awareness matters, but that the strongest mean result emerges only when freshness control is coupled with grounded cross-modal context and conditional fusion.

\paragraph{Dedicated component ablation.}
A separate four-run component sweep provides a more direct, though still non-definitive, decomposition of the final architecture.
Within that sweep, removing lag lowers mean Sharpe from $+0.276$ to $+0.237$, replacing the vector gate with a scalar gate reduces it further to $+0.098$, removing cross-attention yields $+0.130$, and removing web-conditioned gating yields $+0.164$.
These drops are consistent with modality lag, grounded cross-modal context, and vector-valued trust control all contributing useful signal.
One caveat is that the no-gate residual variant reaches $+0.288$, slightly above the full model in that dedicated sweep.
We therefore interpret the ablation conservatively: it supports the importance of lag input, cross-modal grounding, and vector gating, but does not establish a strictly monotonic advantage for every residual design choice at the current sample scale.
The strongest defensible conclusion is thus not that every module is individually optimal, but that the final design instantiates a plausible grounding-plus-trust recipe whose major ingredients receive supporting evidence under controlled sweeps.

\subsection{Modality Lag Analysis}
\label{sec:lag}

To quantify how web-intelligence utility decays with modality age, we construct a \emph{bar-aligned} BTC dataset in which every 15-minute bar is paired with the nearest preceding web snapshot within a 180-minute window (rather than only bars that coincide with a snapshot).
This yields 7{,}295 BTC samples with modality lag uniformly distributed from 0 to 105 minutes (mean 52.3 min, std 34.5 min), enabling direct lag-stratified analysis.

\begin{figure}[tbp]
  \centering
  \includegraphics[width=\linewidth]{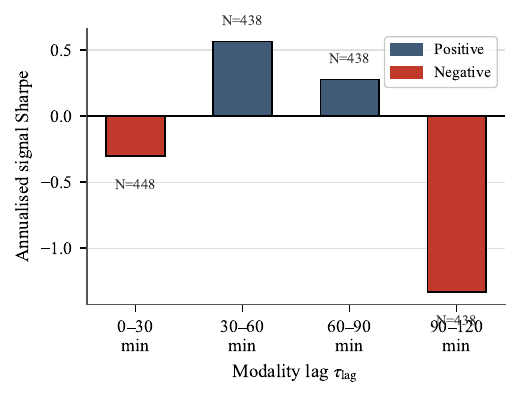}
  \Description{Bar chart showing annualised signal Sharpe by modality lag bin for bar-aligned BTC data. The 0-to-30-minute bin is slightly negative (-0.30). The 30-to-60-minute bin is the highest (+0.57). The 60-to-90-minute bin is lower but still positive (+0.28). The 90-to-120-minute bin is strongly negative (-1.33). Each bin contains approximately 440 samples.}
  \caption{Web-intelligence directional signal Sharpe by modality lag ($\tau_{\mathrm{lag}}$) on the bar-aligned BTC dataset (7{,}295 samples, ${\approx}440$ per bin). The 30--60 min post-publication window provides the strongest positive signal. Very fresh news ($<$30 min) is near-zero, while very stale news ($>$90 min) is strongly negative, indicating a non-monotonic empirical freshness pattern.}
  \label{fig:lag}
\end{figure}

For each sample we compute a directional web-intelligence signal $s = \mathrm{sign}(\texttt{news\_direction\_score})$ and evaluate the annualised Sharpe of a rule that takes a long or short position according to $s$, stratified by $\tau_{\mathrm{lag}}$.
Figure~\ref{fig:lag} shows the result across four 30-minute bins.

The pattern is non-monotonic: the 30--60 min window is the most predictive (Sharpe $+0.57$), while the 0--30 min bin is near-zero ($-0.30$) and the 90--120 min bin strongly negative ($-1.33$).
This is consistent with an empirical \emph{freshness window} in which web intelligence is most useful after a short delay but before it becomes stale.
We interpret Figure~\ref{fig:lag} as a motivating regularity rather than as a definitive market mechanism.
That regularity directly motivates the \CGCMA{} gate design: rather than applying a fixed exponential decay (as in \CGCMAfix{}) or a learned scalar decay (as in \CGCMAlg{}), \CGCMA{} conditions the gate on the full modality representation pair, allowing it to learn a non-monotonic freshness boundary from data.
It also helps explain why simple fixed decay or hard stale filtering can recover only part of the gain in Table~\ref{tab:design}: the useful region is not a single monotone freshness band.
More broadly, the figure shows that asynchronous external context is not merely delayed side information, but a modality whose utility depends on when it is fused.
In practical terms, this suggests that delayed context should be handled through grounded, lag-aware trust rather than through recency alone.

\subsection{Text Diversity as an Independent Bottleneck}
\label{sec:diversity}

\paragraph{Full-year scaling study.}
\label{par:v5}
To test whether ${\geq}500$ samples per fold resolves the multimodal bottleneck, we extend \dataset{} to a full-year BTC corpus spanning January--December 2025 (7{,}335 event-aligned samples, 63 rolling folds).
For January--September 2025, web intelligence is collected with English-only, directionally forced LLM prompts (mean strength $= 0.78$) and merged with the October--December 2025 snapshot data.

Table~\ref{tab:v5_scaling} reports four-run mean$\pm$std results under this expanded protocol.

\begin{table}[tbp]
\caption{Full-year scaling study (v5) on BTC (7{,}335 event-aligned samples; 63 rolling folds; ${\approx}500$ training samples per fold). Entries are mean $\pm$ std across \textbf{4 independent runs}. \CGCMAlg{} is the strongest multimodal variant on the synthetic split.}
\label{tab:v5_scaling}
\centering
\small
\setlength{\tabcolsep}{4pt}
\renewcommand{\arraystretch}{1.05}
\begin{tabular}{lcc}
\toprule
Model & Mean Sharpe & Std. \\
\midrule
\textsc{PriceTx-S} ($d{=}32$) & $\mathbf{+0.124}$ & $\pm 0.056$ \\
\CGCMAlg{} (MiniLM-384 + Mixup) & $+0.021$ & $\pm 0.052$ \\
\bottomrule
\end{tabular}
\vspace{2pt}
\parbox{0.98\columnwidth}{\footnotesize \emph{Note.} Compare with the 111-fold setting (120 samples/fold): \textsc{PriceTx-S} $+0.052{\pm}0.094$; \CGCMAlg{} $+0.158{\pm}0.053$.}
\end{table}

Two findings emerge.
\emph{First}, the price-only model benefits from scale: \textsc{PriceTx-S} improves from $+0.052\pm0.094$ (111-fold) to $+0.124\pm0.056$ (63-fold), with $1.7{\times}$ lower variance.
\emph{Second}, and more strikingly, \CGCMAlg{} \emph{degrades} from $+0.158\pm0.053$ to $+0.021\pm0.052$, reversing the multimodal advantage.
We attribute this reversal to a text signal diversity collapse: January--September 2025 corresponds to the BTC ETF approval and halving cycles, during which LLM-synthesized web intelligence was overwhelmingly bullish ($>$99\% directional signals rated bullish).
When text sentiment is near-monotone, the gate in \CGCMAlg{} cannot extract cross-fold discriminative signal from the text modality, and the fusion model degrades toward a biased predictor.
This finding reframes the scaling hypothesis: the bottleneck for reliable \CGCMA{}-family fusion is not \emph{sample count alone}, but the joint condition of (a) sufficient samples per fold \emph{and} (b) sufficient directional diversity in the text signal.

%% file: sections/08_limitations_ethics.tex
\section{Limitations and Ethics}
\label{sec:limitations}

The present evidence should be interpreted within four boundaries. First, the synthetic web-intelligence text is LLM-generated with an \texttt{as\_of\_date} constraint, so residual parametric leakage cannot be fully ruled out; we therefore use the synthetic split mainly for controlled development and mechanism probing. Second, the real-news corpus is the paper's primary source of external-validity evidence, but it spans only December 2025--March 2026 (about 35 pooled non-overlapping folds), so confidence intervals remain wide and the results should be read as support for problem validity and model plausibility rather than as benchmark-complete or cross-regime evidence. Third, the downstream trading evaluation uses zero transaction costs and zero slippage, so Sharpe should be treated as an offline utility proxy rather than a live-trading projection. Fourth, the task is defined on news-available bars with valid aligned web context rather than on every market bar, making the current conclusions specific to an event-conditioned asynchronous setting. Any deployment in automated trading would require substantial additional validation, risk controls, regulatory compliance, and human oversight.

%% file: sections/09_conclusion.tex
\section{Conclusion}
\label{sec:conclusion}

This paper frames asynchronous alignment as a core multimodal learning problem in which dense primary streams must be fused with sporadic external context whose value depends on when it arrives. Using high-frequency cryptocurrency markets only as a timestamped, high-noise stress test for this event-conditioned setting, \CGCMA{} attains the highest mean downstream Sharpe among the evaluated representative baselines on the current short real-news corpus ($+0.449\pm0.257$ over eight runs). Supporting controls show that this gain is not explained by structured web scalars alone and is not recovered by simple freshness heuristics, while dedicated sweeps support the relevance of lag-aware conditioning, grounded cross-modal context, and vector-valued trust control. The main take-away is simple: delayed external context should be modeled through grounding plus trust, not treated as synchronizable side information. We therefore view \dataset{} as a promising asynchronous evaluation corpus and \CGCMA{} as a problem-driven instantiation of that principle, while leaving broader cross-domain validation to future work.